\documentclass[11pt,letterpaper]{article}
\usepackage{emnlp2016}
\usepackage{times}
\usepackage{latexsym}

\emnlpfinalcopy

\def\emnlppaperid{286}

\title{Product Classification in E-commerce using Distributional Semantics \Thanks{Large part of the work was done when the first author was a Semester Extern with the Ecommerce Company.}}

\author{Vivek Gupta , Harish Karnick\\ Indian Institute of Technology , Kanpur \\
  {\tt \{vgupta,hk\}@cse.iitk.ac.in} \\ \bf { Ashendra Bansal , Pradhuman Jhala} \\ Flipkart Internet Pvt. Ltd., Bangalore\\ {\tt \{ashendra.bansal,pradhuman.jhala\}@flipkart.com}}

\date{}

\begin{document}

\maketitle

\begin{abstract}
Product classification  is the task of automatically predicting a taxonomy path for a product in a  predefined taxonomy hierarchy given a textual product description or title. For efficient product classification we require a suitable representation for a document (the textual description of a product) feature vector and efficient and fast algorithms for prediction.
To address the above challenges, we propose a new distributional semantics representation for document vector formation. We also develop a new two-level ensemble approach utilizing (with respect to the taxonomy tree) path-wise, node-wise and depth-wise classifiers to reduce error in the final product classification task. Our experiments show the effectiveness of the distributional representation and the ensemble approach on data sets from a leading e-commerce platform and achieve improved results on various evaluation metrics compared to earlier approaches.
\end{abstract}

\section{Introduction}

Existing e-commerce platforms  have evolved into large B2C and/or C2C marketplaces having large inventories with millions of products. Products in ecommerce are generally organized into a hierarchical taxonomy of multilevel hierarchical categories. Product classification is an important task in catalog formation and plays a vital role in customer oriented services like search and recommendation and seller oriented services like seller utilities on a seller platform. Product
classification is a hierarchical classification problem and presents the following challenges: a)  a large number of categories have data that is extremely sparse with a skewed long tailed distribution, b) a hierarchical taxonomy imposes constraints on activation of labels. If a child label is active then it is necessary for a parent label to be active, c) for practical use the prediction should happen in real time - ideally within few milli-seconds.

Traditionally, documents have been represented as a weighted bag-of-words (BoW) or tf-idf feature vector, which contains weighted information about the presence or absence of words in a document by using a fixed length vector. Words that define the semantic content of a document are expected to be given higher weight. While tf-idf and BoW representations perform well for simple multi-class classification tasks, they generally do not do as well for more complex tasks because the BoW representation ignores word ordering and polysemy, is extremely sparse and high dimensional and does not encode word meaning.


Such disadvantages have motivated continuous, low-dimensional, non-sparse distributional representations. A word is encoded as a vector in
a low dimension vector space typically $\mathcal{R}^{100}$ to $\mathcal{R}^{300}$. The vector encodes local context and therefore is sensitive to local word order and captures word meaning to some extent. It relies on the `Distributional Hypothesis'\cite{harris54} i.e. \textit{“Similar words occur in similar contexts”}. Similarity between two words can be calculated via cosine distance between their vector representations.
Le and Mikolov \cite{le2014distributed} proposed paragraph vectors, which use global context together with local context to represent documents. But paragraph vectors suffer from the following problems:
a) current techniques embed paragraph vectors in the same space (dimension) as word vectors although a paragraph can consist of words belonging to multiple topics (senses), b) current techniques also ignore the importance and distinctiveness of words across documents. They assume all words contribute equally both quantitatively (weight) and qualitatively (meaning).

In this paper we describe a new compositional technique for formation of document vectors from semantically enriched word vectors to address the above problems. Further, to capture importance, weight and distinctiveness of words across documents we use a graded weights approach, inspired by the work of Mukerjee et al. \cite{pranjal2015weighted}, for our compositional model. We also propose a new two-level approach  for product classification which uses an ensemble of classifiers for label paths, node labels and depth-wise labels (with respect to the taxonomy) to decrease classification error . Our new ensemble technique efficiently exploits the catalog hierarchy and achieves improved results in top $K$ taxonomy path prediction. We show the effectiveness of the new representation and classification approach for product classification of two e-commerce data-sets containing book and non-book descriptions. 

\section{Related Work}
\subsection{Distributional Semantic Word Representation}

The distributional word embedding method was first introduced by Bengio et al. as the Neural Probabilistic Language Model \cite{bengio2003neural}.   
Later, Mikolov et al. \cite{mikolov2013efficient} proposed a simple log-linear model which considerably reduced training time - Word2Vec  Continuous Bag-of-Words (CBoW) model and Skip-Gram with Negative Sampling (SGNS) model. Figure 1 shows the architecture for CBoW (Left) and Skip-Gram (Right). 






Later Glove \cite{glove2014distributed} a log-bilinear model with a weighted least-squares objective  was proposed which uses the statistical ratio of global word-word co-occurrences in the corpus for training word vectors. The word vectors learned using the skip-gram model are known to encode many linear linguistic regularities and patterns \cite{levy2014regularities}.



While the above methods look very different they implicitly factorize a shifted positive point-wise mutual information matrix (PPMI) with tuned hyper parameters as shown by Levy and Goldberg \cite{NIPS2014_5477}. Some variants incorporate ordering information in context words to capture syntactic information by replacing summation of context word vectors with concatenation during training \cite{wang2015} of CBoW and SGNS models. 


\subsection{Distributional Paragraph Representation}
Most models for learning distributed representations for long text such as phrases, sentences or documents that try to capture semantic composition do not go beyond simple weighted average of word vectors. This approach is analogous to a bag-of-words approach and neglects word order while representing documents. Socher et al. \cite{socher2013recursive} propose a recursive tensor neural network where the dependency parse-tree of the sentence is used to compose word vectors in a bottom-up approach to represent sentences or phrases. This approach considers syntactic dependencies but cannot go beyond sentences as it depends on parsing.

Mikolov proposed a distributional paragraph vector framework called  paragraph vectors which are trained in a manner similar to word vectors. He proposed two types of models called \textit{Distributed Memory Model Paragraph Vectors (PV-DM)} \cite{le2014distributed} 
and \textit{Distributed BoWs paragraph vectors (PV-DBoW)} \cite{le2014distributed}. In PV-DM the model is trained to predict the center word using context words in a small window and the paragraph vector \cite{le2014distributed}. Here context words to be predicted are represented by $w_{t-k}$,....,$w_{t+k}$  and the document vector is represented by $D_{i}$. In PV-DBoW the paragraph vector is trained to predict context words directly. Figure 2 shows the network architecture for PV-DM(L) and PV-DBoW(R).

The paragraph vector presumably represents the global semantic meaning of the paragraph and also incorporates properties of word vectors i.e. meanings of the words used. A paragraph vector exhibits close resemblance to an n-gram model with a large $n$. This property is crucial because the n-gram model preserves a lot of information in a sentence (and the paragraph) and is sensitive to word order. This model mostly performs better than the BoW models which usually create a very high-dimensional representation leading to poorer generalization.

\begin{figure}
\centering
\includegraphics[scale=0.5]{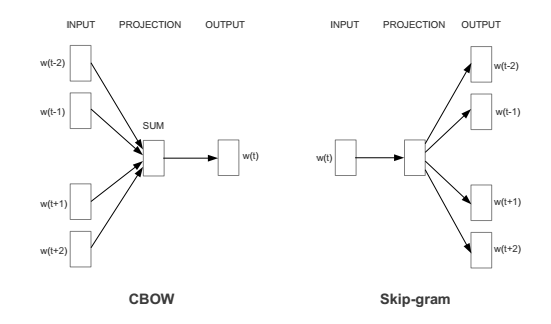}
\caption{Neural Network Architecture for CBoW and Skip Gram Model}
\label{figure:1}
\end{figure}

\begin{figure}
\centering
\includegraphics[scale=0.8]{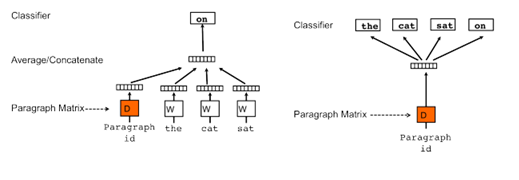}
\caption{Neural Network Architecture for Distributed Memory version of Paragraph Vector (PV-DM) and Distributed BoWs version of paragraph vectors (PV-DBoW)}
\label{figure:3}
\end{figure}


\subsection{Problem with Paragraph Vectors}
Paragraph vectors obtained from PV-DM and PV-DBoW are shared across context words generated from the same paragraph but not across paragraphs. On the other hand a word is shared across paragraphs. Paragraph vectors are also represented in the same space (dimension) as word vectors though a paragraph can contain words belonging to multiple topics (senses). The formulation for paragraph vectors ignores the importance and distinctiveness of a word across documents i.e. assumes all words contribute equally both quantitatively (weight wise) and qualitatively (meaning wise). Quantitatively, only binary weights i.e. 0 weight for stop-words and non-zero weight for others are used. Intuitively, one would expect the paragraph vector to be embedded in a larger and enriched space. 

\subsection{Hierarchical Product Categorization}
\label{treebased}
\label{treebased}
Most methods for hierarchical classification follow a \textit{“gates-and-experts”} method which have a two level classifier. The high-level classifier serves as a ``gate'' to a lower level classifier called the ``expert'' \cite{Shen2011categorization}. The basic idea is to decompose the problem into two models, the first model is simple and does coarse-grained classification while the second model is more complex and does more fine-grained classification. The coarse-grained classification deals with a huge number of examples while the fine-grained distinction is learned within a subtree under every top level category with better feature generation and classification algorithms and deals with fewer categories. 

Kumar et al. \cite{Kumar2002}, proposed an approach that learnt a tree structure over the set of classes. They used a clustering algorithm based on Fisher’s discriminant that clustered training examples into mutually exclusive groups inducing a partitioning on the classes. As a result the prediction by this method is faster but the training process is slow as it involves solving many clustering problems. 

Later, Xue et al. \cite{Xue2008deepclassification} suggested an interesting two stage strategy called \textit{``deep classification''}. The first stage (search) groups documents in the training set that are similar to a given document. In the second stage (classification) a classifier is trained on these classes and used to classify the document. In this approach a specific classifier is trained for each document making the algorithm computationally
inefficient.

For large scale classification Bengio et al. \cite{NIPS2010_4027} use the confusion matrix for estimating class similarity instead of clustering data samples. Two classes are assumed to be similar if they are often confused by a classifier. Spectral clustering, where the edges of the similarity graph are weighted by class confusion probabilities, is used to group similar classes together.

Shen and Ruvini \cite{Shen2012categorization} \cite{Shen2011categorization} extend the previous approach by using a mixture of simple and complex classifiers for separating confused classes rather then spectral clustering methods which has faster training times. They approximate the similarity of two classes by the probability that the classifier incorrectly predicts one of the categories when the correct label is the other category. Graph algorithms are used  to generate connected groups from estimated confusion probabilities. They represent the relationship among classes using an undirected graph $G = (V,E)$, where the set of vertices $V$ is the set of all classes and $E$ is the set of all edges. Two vertices's are connected by an edge if the confusion probability $Conf(c_{1}, c_{2})$ is greater than a given threshold $\alpha$ \cite{Shen2012categorization}.

Other simple approaches like flat classification and top down classification are intractable due to the large number of classes and give poor results due to error propagation as described in \cite{Shen2012categorization}.  

\section{Graded Weighted Bag of Word Vectors}

We propose a new method to form a composite document vector using word vectors i.e. distributional meaning and tf-idf and call it a Graded Weighted Bag of Words Vector (gwBoWV). gwBoWV is 
inspired from the computer vision literature where we use a Bag of Visual words to form feature vectors. gwBoWV is calculated as follows:

\begin{enumerate}
\item Each document is represented in a lower dimensional space 
$D = K*d + K$, where $K$ represents number of semantic clusters and $d$ is the dimension of the word-vectors.
\item Each document is also concatenated with inverse cluster frequency(icf) values which is calculated using idf values of words present in the document. 
\end{enumerate}

Idf values from the training corpus are directly used for the test corpus for weighting. Word vectors are first separated into a pre-defined number of semantic clusters using a suitable clustering algorithm (e.g. k-means). For each document we add the word-vectors of each word in the document belonging
to a cluster to form a cluster vector. We finally concatenate the cluster vector and the icf for each of the $K$ clusters to obtain the document vector. Algorithm \ref{gwBoWV} describes this in more detail.
\begin{algorithm}
    \SetAlgoNoLine
    \KwData{Documents $D_{n}$, n = 1 $\ldots$ N}
    \KwResult{Document vectors $\vec{gwBoWV_{D_{n}}}$, n = 1 $\ldots$  N}
    Train SGNS model to obtain word vector representation ($wv_{n}$) using all document $D_{n},\,n = 1..N$\;
    Calculate idf values for all words: $idf(w_{j}),\,j = 1..|V|$ \tcc*{$|V|$ is vocabulary size}
    Use K-means algorithm for clustering all words in $V$ using their word-vectors into K clusters\;
    \For{ $i\in (1..N)$}{
    Initialize cluster vector $\vec{cv_{k}}$ = $\vec{0},\,k = 1..K$\;
    Initialize cluster frequency $icf_{k} = 0,\,k = 1..K$\;
    \While{not at end of document $D_{i}$}{
        read current word $w_{j}$ and obtain wordvec $\vec{wv_{j}}$\;
        obtain cluster index $k$ = $idx(\vec{wv_{j}})$ for wordvec $\vec{wv_{j}}$\;
        update cluster vector $\vec{cv_{k}}$ $+=$ $\vec{wv_{j}}$\;
        update cluster frequency $icf_{k}$ $+=$ $idf(w_{j})$\;
    }
    obtain $\vec{gwBoWV_{D_{i}}}= \bigoplus_{k=1}^K \vec{cv_{k}}\oplus icf_{k}$ \tcc*{$\oplus$ is concatenation}
}

\caption{Graded Weighted Bag of Word Vectors}
\label{gwBoWV}
\end{algorithm}


Since semantically different vectors are in separate clusters we avoid averaging of semantically different words during Bag of Words Vector formation.
Incorporation of idf values captures the weight of each cluster vector which tries to model the importance and distinctiveness of words across documents. 


\section{Ensemble of Multitype Predictors}
\label{section:ensemble_theory}
We propose a two level ensemble technique to combine multiple classifiers predicting product paths, node labels and  depth-wise labels respectively. We  construct an ensemble of multi-type features for categorization inspired by the recent work of Zornitsa et. el. from Yahoo Labs \cite{yahoo2015}. Below are the details of each classifier used at level one:

\begin{itemize}
    \item {\em Path-Wise Prediction Classifier}: 
    We take each possible \textit{path} in the catalog taxonomy tree, from leaf node to root node,  as a possible class label and train a classifier ($PP$) using these labels.
    \item {\em Node-Wise Prediction Classifier}:
    We take each possible \textit{node} in the catalog taxonomy tree as a possible prediction class and train a classifier ($NP$) using these class labels.
    \item {\em Depth-Wise Node Prediction Classifiers}:
   We train multiple classifiers ($DNP_{i}$) one for each depth level of the
taxonomy tree. Each possible \textit{node} in the catalog taxonomy tree at that depth is a possible class label. All data samples which have a potential node at depth $k$, in addition 10\% samples of data points which have no node at depth $k$ (sample of data point whose path ended before depth $k$) are used for training.
\end{itemize}

We use the output probabilities of these classifiers at level one ($PP$, $NP$, $DNP_{i}$) as a feature vector and train a classifer (level two) after some dimensionality reduction. 

The increase in training time can be reduced by  training all level one classifiers in parallel. The algorithm for training the ensemble is described in Algorithm \ref{train_ensemble}. The testing algorithm is similar to training and described in supplementary section  \ref{test_ensemble}.

\label{tlba}
\begin{algorithm}
    \SetAlgoNoLine
    \KwData{Catalog Taxonomy Tree (T) of depth K and training data $D$ = (d, $p_{d}$) where $d$ is the product description and $p_{d}$ is the taxonomy path label.}
    \KwResult{ Set of level  one Classifiers C = \{{$PP, NP, DNP_{1},\ldots, DNP_{K}$\}} and level two classifier $FPP$.}
    Obtain $\vec{gwBoWV}_{d}$ features for each product description d \;
    Train Path-Wise Prediction Classifier ($PP$) with possible classes as product taxonomy paths ($p_{d}$)\;
    Train Node-Wise Prediction Classifier ($NP$) with possible classes as nodes in taxonomy path i.e. ($n_{d}$). Here each description will have multiple node labels. \\
    \For{k $\in$ ($1\ldots K$)}{
    Train Depth-Wise Node Classifier for depth $k$ ($DNP_{k}$) with labels as nodes at depth $k$ i.e. ($n_{k}$)
    } 
    Obtain output probabilities $\vec{P}_{X}$ over all classes for each level one classifier $X$ i.e. $\vec{P}_{PP}$, $\vec{P}_{NP}$ and $\vec{P}_{DNP_{k}},\,k=1..K$.\;
    Obtain feature vector $\vec{FV}_{d}$ for each description as:
    \begin{equation}
    \label{fv}
        \vec{FV}_{d} = \vec{gwBoWV}_{d} \oplus \vec{P}_{PP} \oplus \vec{P}_{NP} \bigoplus_{k=1}^K \vec{P}_{DNP_{k}} 
    \end{equation} 
    \tcc{$\bigoplus$ is the concatenation operation}
    Reduce feature dimension ($\vec{RFV}_{d}$) using suitable supervised feature selection technique based on mutual information criteria\;
    Train Final Path-Wise Prediction Classifier ($\vec{FPP}_{d}$) using $RFV_{d}$ as feature vector and possible class labels as product taxonomy paths ($p_{d}$) 
\caption{Training Two Level Boosting Approach}
\label{train_ensemble}
\end{algorithm}

\section{Dataset}
\begin{table}
\begin{tabular}{|l|l|l|}
\hline
\bf Level              & \bf \#Categories            & \bf \%Data Samples \\ \hline
1   & 21       & 34.9\%  \\ \hline
2   & 278        & 22.64\%  \\ \hline
3   & 1163      &  25.7\%  \\ \hline
4   & 970       & 12.9\%  \\ \hline
5   & 425         & 3.85\%  \\ \hline
6   & 18     & 0.10\%  \\ \hline
\end{tabular}
\caption{Percentage of Book Data ending at each  depth level of the book taxonomy hierarchy which had a maximum depth of 6.}
\label{Table:5.1}
\end{table}

We use seller product descriptions and title samples from a leading e-commerce site for experimentation\footnote{This data is proprietary to the e-commerce Company.}. The data set had two product taxonomies:{\em  non-book} and {\em book}. Non-book data is more discriminative with average description + title length of around 10 to 15 words, whereas book descriptions have an average length greater than 200 words. To give more importance to the title we generally weight it three times the description value. The distribution of items over leaf categories (verticals) exhibits high skewness and heavy tailed nature and suffers from sparseness as shown in Figure 3. We use random forest and k nearest neighbor as base classifiers as they are less affected by data skewness

We have removed data samples with multiple paths to simplify the problem to single path prediction. Overall, we have 0.16 million training and 0.11 million testing samples for book data and 
0.5 million training and 0.25 million testing samples for non-book data. Since the taxonomy evolved over time all category nodes are not semantically mutually exclusive. Some ambiguous leaf categories are even meta categories. We handle this by giving a unique id to every node in the category tree of book-data. Furthermore, there are also category paths with different categories at the top and similar categories at the leaf nodes i.e. reduplication of the same path with synonymous labels. 

The quality of the descriptions and titles also varies a lot. There are titles and descriptions that do not contain enough information to decide an unique appropriate category. There were labels like \textit{Others} and \textit{General} at various depths in the taxonomy tree which carry no specific semantic meaning. Also, descriptions with the special label  `wrong procurement' are removed manually for consistency.

\begin{figure}
\centering
\includegraphics[scale=0.2]{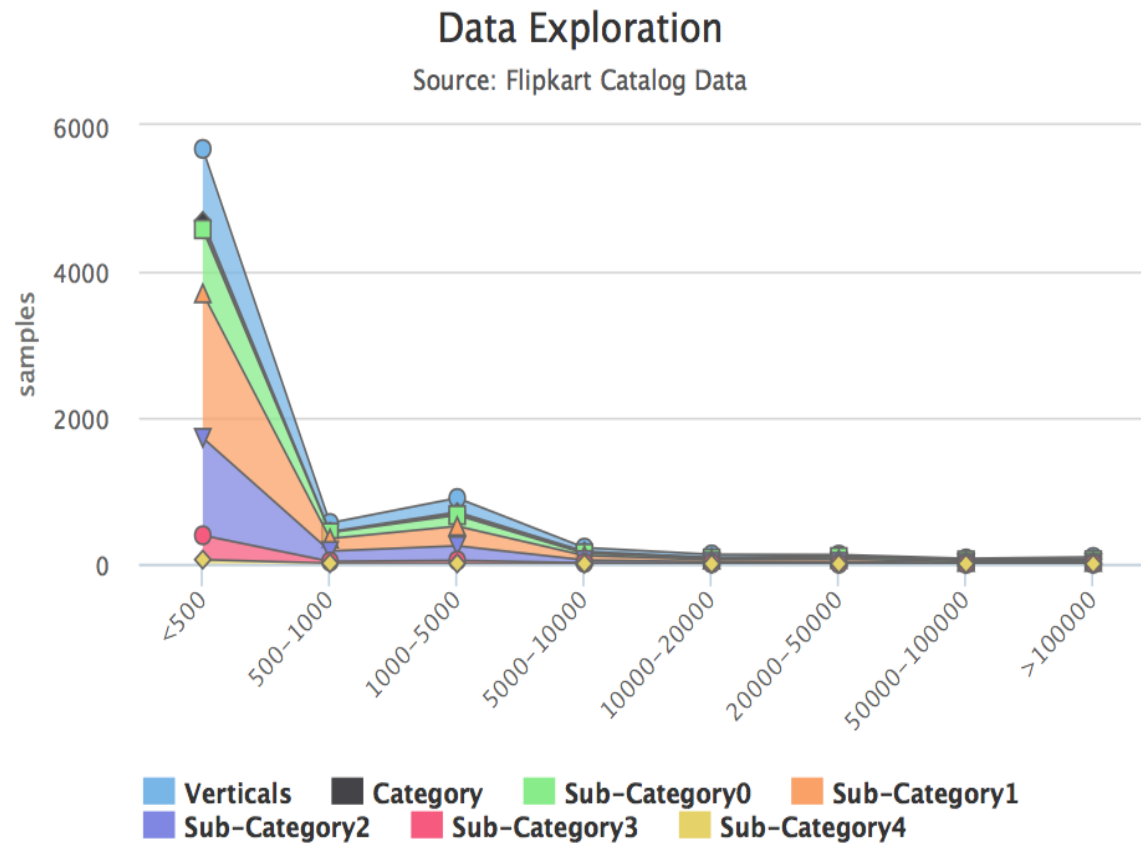}
\caption{ Figure shows distribution of items over sub-categories and leaf category (verticals) for non-book dataset}
\label{figure:7}
\end{figure}

\begin{figure}
\centering
\includegraphics[scale=0.2]{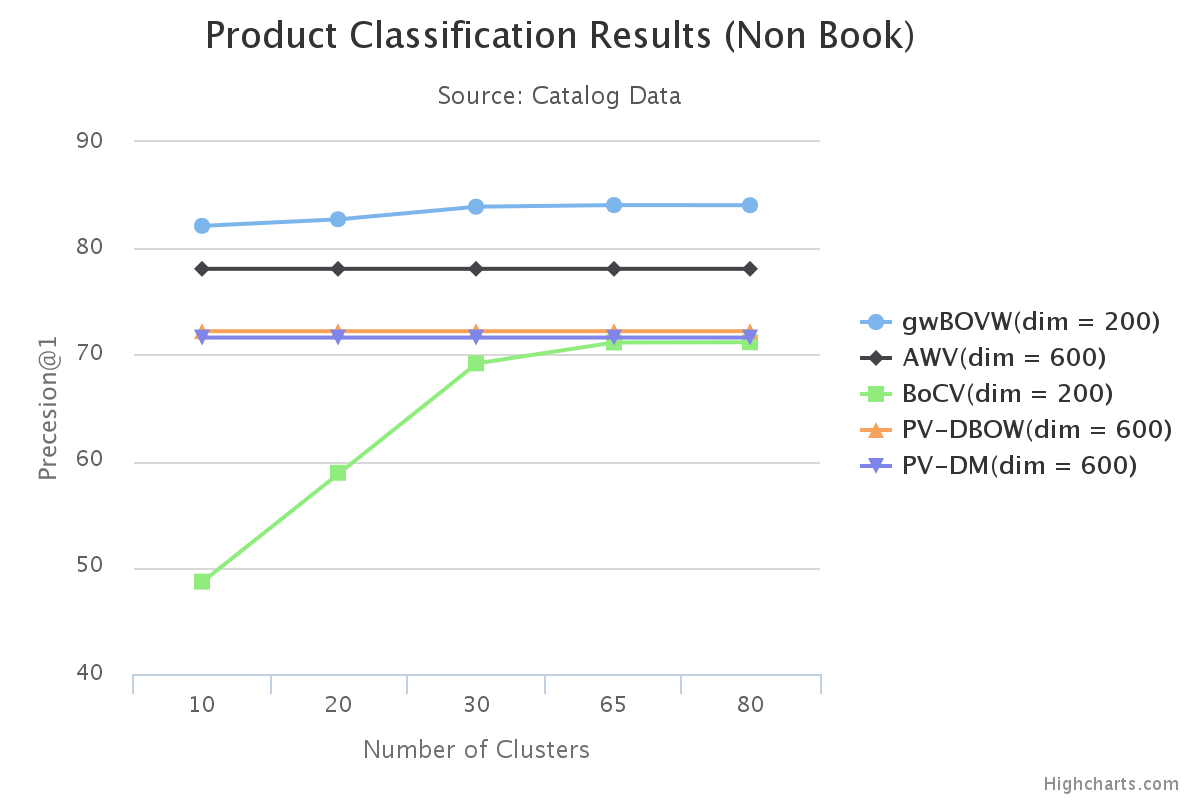}
\caption{Comparison of prediction accuracy for path prediction using different methods for document vector generation.}
\label{figure:8}
\end{figure}

\begin{table}
\centering
\begin{tabular}{|l|l|l|}
\hline
\bf Level              & \bf \#Categories            & \bf \%Data Samples \\ \hline
1   & 21       & 34.9\%  \\ \hline
2   & 278        & 22.64\%  \\ \hline
3   & 1163      &  25.7\%  \\ \hline
4   & 970       & 12.9\%  \\ \hline
5   & 425         & 3.85\%  \\ \hline
6   & 18     & 0.10\%  \\ \hline
\end{tabular}
\caption{Percentage of Book Data ending at each  depth level of the book taxonomy hierarchy which had a maximum depth of 6.}
\label{Table:5.1}
\end{table}



The quality of the descriptions and titles also varies a lot. There are titles and descriptions that do not contain enough information to decide an unique appropriate category. There were labels like \textit{Others} and \textit{General} at various depths in the taxonomy tree which carry no specific semantic meaning. Also, descriptions with the special label  `wrong procurement' are removed manually for consistency.

\section{Results}

The classification system is evaluated using the usual precision metric defined as fraction of products from test data for which the classifier predicts correct taxonomy paths. Since there are multiple similar paths in the data set predicting a single path is not appropriate. One solution is to predict more than one path or better a ranked list of of 3 to 6 paths with  predicted label coverage matching labels in the true path. The ranking is obtained using the confidence score of the predictor. We also calculate the confidence score of the correct prediction path by using the $k$ (3 to 6) confidence scores of the individual predicted paths. For the purpose of measuring accuracy when more than one path is predicted, the classifier result is counted as correct when the correct class (i.e. path assigned by seller) is one of the returned class (paths). Thus we calculated Top 1, Top 3 and Top 6 prediction accuracy when 1, 3 and 6 paths are predicted respectively.
\subsection{Non-Book Data Result}
We also compare our results with document vectors formed by averaging word-vectors of words in the document i.e. Average Word Vectors (AWV), Distributed Bag of Words version of Paragraph Vector by Mikolov (PV-DBoW), Frequency Histogram of word distribution in Word-Clusters i.e. Bag of Cluster Vector (BoCV). We keep the classifier (random forest with 20 trees) common for all document vector representations. We compare performance with respect to number of clusters, word-vector dimension, document vector dimension and vocabulary dimension (tf-idf) for various models.


Figure 4 shows results for a random forest (20 trees) on various classifiers trained by various methods on 0.2 million training and 0.2 million testing samples with 3589 classes. It compares our approach gwBoWV with PV-DBoW and PV-DM models with varying word vector dimension and number of clusters. The dimension of word vector for gwBoVW and BoCV is 200. Note AWV , PV-DM and PV-DBoW are independent of cluster number and have dimension 600. Clearly gwBoWV performs much better than other methods especially PV-DBoW and PV-DM.

Table \ref{Table:5.5} Shows the effect of varying cluster numbers on accuracy for Non Book Data for 2 lakh training and 2 lakh testing using 200 dimension word vector.

\begin{table}[H]
\centering
\begin{tabular}{|l|l|l|}
\hline
\bf\#Cluster  & \bf CBoW & \bf SGNS \\ \hline
10  & 81.35\% & 82.12\% \\ \hline
20 & 82.29\%  & 82.74\%\\ \hline
50 &  83.66\%  & 83.92\% \\ \hline
65 &  83.85\%   & 84.07\%\\ \hline
80 &  83.91\%  & 84.06\%\\ \hline
100 & 84.40\% & 84.80\%\\ \hline
\end{tabular}
\caption{\textit{Result of classification on varying Cluster Numbers for fixed word vector size 200 for Non Book Data for CBow and SGNS architecture \#Train Sample = 0.2 million , \#Test Sample = 0.2 million}}
\label{Table:5.5}
\end{table}

We use the notation given below to define our evaluation metrics for Top K path prediction :

\begin{itemize}

\setlength\itemsep{0.5em}

\item $\tau^*$ represents the true path for a product description.

\item $\tau_{i}$ represents the $i_{th}$ predicted path by our algorithm , where i $\in$ $\{1,2 \ldots K \}$.

\item $\mathbb{T}^{*}$ represent the nodes in true path ${\tau^*}$.

\item $\mathbb{T}^{i}$ represents the nodes in $i_{th}$ predicted path ${\tau_{i}}$, where i $\in$ $\{1,2 \ldots K \}$.

\item p($\tau^*$) represents the probability predicted by our algorithm for true path ${\tau^*}$. p($\tau^*$) = 0 if $\tau^*$ $\notin$  $\{\tau_{1}, \tau_{2} \ldots \tau_{K}\}$ 

\item p($\tau_{i}$) represents the probability of $i_{th}$ predicted path by our algorithm, here i $\in$ $\{1,2 \ldots K \}$.

\end{itemize}

We use four evaluation metrics to measure performance for the top $k$ predictions as described below:

\begin{enumerate}
\item Prob Precision $@$ K (PP) : $PP@K$ = p($\tau^{*}$) / (p($\tau_{1}$) +  p($\tau_{2}$) + $\ldots$ +  p($\tau_{K}$)).

\item Count Precision $@$ K (CP) : $CP@k$ = 1 if $\tau^{*}$ $\in$ $\{ \tau_{1} , \tau_{2} \ldots \tau_{K} \}$ else $CP@K$ =  0.

\item Label Recall $@$ K (LR) : 
$LR@k$ = $\|\mathbb{T}^{*} \cap  ( \cup{_{1}^{K}} 
\mathbb{T}^{i})\| / \|\mathbb{T}^{*}\|$.Here $\|$S$\|$ represent number of elements in set S.

\item Label Correlation $@$ k (LC) : 
$LC@k$ =  $\|\cap{_{1}^{K}} 
\mathbb{T}^{i}\| / \|\cup{_{1}^{K}}
\mathbb{T}^{i}\|$ . Here $\|$S$\|$ represent number of elements in set S.
\end{enumerate}

Table \ref{Table:5.12} shows the results on all evaluation metrics with varying word-vec dimension and clusters. Table \ref{Table:5.15} shows results of top 6 paths prediction for tfidf baseline with varying dimension.

\begin{table}[H]
\centering
\begin{tabular}{|l|l|l|l|l|l|}
\hline
\bf \#Clus, \bf \#Dim & \bf \%PP & \bf \%CP & \bf \%LR & \bf \%LC\\ \hline
40, 50 &  82.07 & 96.43 & 98.27  & 34.50 \ \\ \hline
40, 100 & \bf 83.18 & 96.67 & 98.39  & \bf 34.91 \\ \hline
100, 50 & 82.05 & 96.40 & 98.26  & 34.41 \\ \hline
100,100 & 83.13 & \bf 96.75 & \bf 98.42  & 34.88\\ \hline
\end{tabular}
\caption{\textit{Result for top 6 paths predicted for multiple Bag of Word Vectors with varying dimension and number of clusters with weighting on Non-Book Data with \#Train Samples = 0.50 million, \#Test  Samples  = 0.35 million.}}
\label{Table:5.12}
\end{table}

\begin{table}[H]
\centering
\begin{tabular}{|l|l|l|l|l|}
\hline
\bf \#Dim & \bf \%PP & \bf \%CP & \bf \%LR & \bf \%LC\\ \hline
 2000 & 81.10 & 94.04 & 96.85  & 35.37  \\ \hline
 4000 & \bf 82.74 &\bf  94.78 &\bf  97.33  &\bf  35.61 \\ \hline
\end{tabular}
\caption{\textit{Result of top 6 paths prediction for tfidf with varying dimension on  Non Book Data \#Train Samples = 0.50 million, \#Test  Samples  = 0.35 million.}}
\label{Table:5.15}
\end{table}

\subsection{Book Data Result}
Book data is harder to classify. There are more cases of improper paths and labels in the taxonomy and hence we had to do a lot of pre-processing. Around 51\% of the books did not have labels at all and 15\%  books were given extremely ambiguous labels like `general' and `others'. To maintain consistency we prune the above 66\% data samples and work with the  remaining 44\% i.e. 0.37 million samples.

To handle improper labels and ambiguity in the taxonomy we use multiple classifiers one predicting path (or leaf) label, another predicting node labels and multiple classifiers, one at each depth level of the taxonomy tree, that predict node labels at that level. In depth-wise node classification we also introduce the `none' label to denote missing labels at a particular level i.e. for paths that end at earlier levels. However we only take a random strata sample for this `none' label. 

\subsection{Ensemble Classification}
We use the ensemble of multi-type predictors as describe in Section \ref{section:ensemble_theory} for final classification. For dimensionality reduction we use feature selection methods based on mutual information criteria (ANOVA F-value i.e. analysis of variance). We obtain improved results for all four evaluation metrics with the new ensemble technique as shown in Table~\ref{Table:5.21}for Book Data.The list below says how the first column in Table 5 should be interpreted
\begin{itemize}
    \item tf-idf (A-2-C): term frequency and inverse document frequency feature with \#A top 1, 2 gram words and \#C random forest trees
    \item path-1 (A-C): path prediction model without ensemble, trained with gwBoWV with \#A (cluster*wordvec dimension) using C trees
    \item Dep (A+B-C): trained with gwBoWV with A features, B represents size of out-probability vectors (\#total nodes) for all depths using depth classifier of level 1 using \#C trees.
    \item node (A+B): trained with gwBoWV with A features, B represents size of output probability vector(\#total nodes) by level 1 node classifier using \#C tree.
    \item comb-2(A): level two combined ensemble classifier with A reduced features (original features 21706).
\end{itemize}


\begin{table}[H]
\centering
\begin{tabular}{|l|l|l|l|l|}
\hline
\bf Method & \bf PP & \bf CP & \bf LR & \bf LC\\ \hline
tfidf(4000-2-20) & 41.33 & 75.00 & 86.86 & 22.14\\ \hline
tfidf(8000-2-20) & 41.39 & 74.95 & 86.83 & 22.16\\ \hline
tfidf(10000-2-20) & 41.39 & 74.96 & 86.85 & 22.18\\ \hline
path-1(4100-15) & 39.86 & 74.17 & 86.37 & 22.19\\ \hline
path-1(8080-20) & 41.08 & 74.83 & 86.60 & 22.19\\ \hline
Dep(5100+2875-20) & 41.54 & 75.34 & 87.08 & 22.47\\ \hline
node(4100+2810) & 41.54 & 74.68 & 86.65 & 22.34\\ \hline
comb-2(8000) & 45.64 & \bf 77.26 & \bf 88.86 & 24.57\\ \hline
comb-2(6000) & \bf 46.68 & 75.74 & 87.67 & \bf 25.08\\ \hline
comb-2(10000) & 42.82 & 75.83 & 87.62 & 23.08\\ \hline
\end{tabular}
\caption{\textit{Results from various approaches for Top 6 predictions for Book Data}}
\label{Table:5.21}
\end{table}


\section{Conclusions}

We presented a novel compositional technique using embedded word vectors to form appropriate document vectors.  Further, to capture importance, weight and distinctiveness of words across documents we used a graded weighting approach for composition based on recent work by Mukerjee et. el. \cite{pranjal2015weighted} where instead of weighting we weight using cluster frequency. Our document vectors are embedded in a vector space different from
the word embedding vector space. This document vector space is higher dimensional and tries to encode the intuition that a document has more
topics or senses than a word.  


We also developed a new technique which uses an ensemble of multiple classifiers that predicts label paths, node labels and depth-wise labels to decrease classification error. We tested our method on data sets from a leading e-commerce platform and show improved performance when compared with competing approaches.

\section{Future Work}

Instead of using k-means we can use the Chinese Restaurant Process. Extending the gwBOVW approach to learn supervised class document vectors that consider the label in some fashion during embedding.




\bibliography{emnlp2016}

\begin{thebibliography}{}

\bibitem[\protect\citename{Bengio \bgroup et al.\egroup
  }2003]{bengio2003neural}
Yoshua Bengio, R{\'e}jean Ducharme, Pascal Vincent, and Christian Janvin.
\newblock 2003.
\newblock A neural probabilistic language model.
\newblock {\em The Journal of Machine Learning Research}, 3:1137--1155.

\bibitem[\protect\citename{Bengio \bgroup et al.\egroup }2010]{NIPS2010_4027}
Samy Bengio, Jason Weston, and David Grangier.
\newblock 2010.
\newblock Label embedding trees for large multi-class tasks.
\newblock In {\em Advances in Neural Information Processing Systems 23}, pages
  163--171. Curran Associates, Inc.

\bibitem[\protect\citename{Bi and Kwok}2011]{Bi11multi-labelclassification}
Wei Bi and James~T. Kwok.
\newblock 2011.
\newblock Multi-label classification on tree- and dag-structured hierarchies.
\newblock In {\em In ICML}.

\bibitem[\protect\citename{Harris}1954]{harris54}
Zellig Harris.
\newblock 1954.
\newblock Distributional structure.
\newblock {\em Word}, 10:146--162.

\bibitem[\protect\citename{Jeffrey~Pennington}2014]{glove2014distributed}
Christopher D.~Manning Jeffrey~Pennington, Richard~Socher.
\newblock 2014.
\newblock Glove: Global vectors for word representation.
\newblock In {\em Empirical Methods in Natural Language Processing (EMNLP)},
  pages 1532--1543. ACL.

\bibitem[\protect\citename{Kozareva}2015]{yahoo2015}
Zornitsa Kozareva.
\newblock 2015.
\newblock Everyone likes shopping! multi-class product categorization for
  e-commerce.
\newblock {\em Human Language Technologies: The 2015 Annual Conference of the
  North American Chapter of the ACL}, pages 1329--1333.

\bibitem[\protect\citename{Kumar \bgroup et al.\egroup }2002]{Kumar2002}
Shailesh Kumar, Joydeep Ghosh, and M.~Melba Crawford.
\newblock 2002.
\newblock Hierarchical fusion of multiple classifiers for hyperspectral data
  analysis.
\newblock {\em Pattern Analysis and Applications}, 5:210--220.

\bibitem[\protect\citename{Le and Mikolov}2014]{le2014distributed}
Quoc~V Le and Tomas Mikolov.
\newblock 2014.
\newblock Distributed representations of sentences and documents.
\newblock {\em arXiv preprint arXiv:1405.4053}.

\bibitem[\protect\citename{Levy and Goldberg}2014a]{levy2014dependencybased}
Omer Levy and Yoav Goldberg.
\newblock 2014a.
\newblock Dependencybased word embeddings.
\newblock {\em Proceedings of the 52nd Annual Meeting of the Association for
  Computational Linguistics}, 2:302--308.

\bibitem[\protect\citename{Levy and Goldberg}2014b]{levy2014regularities}
Omer Levy and Yoav Goldberg.
\newblock 2014b.
\newblock Linguistic regularities in sparse and explicit word representations.
\newblock In {\em Proceedings of the Eighteenth Conference on Computational
  Natural Language Learning}, pages 171--180. Association for Computational
  Linguistics.

\bibitem[\protect\citename{Levy and Goldberg}2014c]{NIPS2014_5477}
Omer Levy and Yoav Goldberg.
\newblock 2014c.
\newblock Neural word embedding as implicit matrix factorization.
\newblock In {\em Advances in Neural Information Processing Systems 27}, pages
  2177--2185.

\bibitem[\protect\citename{Mikolov \bgroup et al.\egroup
  }2013a]{mikolov2013efficient}
Tomas Mikolov, Kai Chen, Greg Corrado, and Jeffrey Dean.
\newblock 2013a.
\newblock Efficient estimation of word representations in vector space.
\newblock {\em arXiv preprint arXiv:1301.3781}.

\bibitem[\protect\citename{Mikolov \bgroup et al.\egroup
  }2013b]{mikolov2013distributed}
Tomas Mikolov, Ilya Sutskever, Kai Chen, Greg~S Corrado, and Jeff Dean.
\newblock 2013b.
\newblock Distributed representations of words and phrases and their
  compositionality.
\newblock In {\em Advances in Neural Information Processing Systems}, pages
  3111--3119.

\bibitem[\protect\citename{Pranjal~Singh}2015]{pranjal2015weighted}
Amitabha~Mukerjee Pranjal~Singh.
\newblock 2015.
\newblock Words are not equal: Graded weighting model for building composite
  document vectors.
\newblock In {\em Proceedings of the twelfth International Conference on
  Natural Language Processing (ICON-2015)}. BSP Books Pvt. Ltd.

\bibitem[\protect\citename{Shen \bgroup et al.\egroup
  }2011]{Shen2011categorization}
Dan Shen, Jean~David Ruvini, Manas Somaiya, and Neel Sundaresan.
\newblock 2011.
\newblock Item categorization in the e-commerce domain.
\newblock In {\em Proceedings of the 20th ACM International Conference on
  Information and Knowledge Management}, CIKM '11, pages 1921--1924.

\bibitem[\protect\citename{Shen \bgroup et al.\egroup
  }2012]{Shen2012categorization}
Dan Shen, Jean-David Ruvini, and Badrul Sarwar.
\newblock 2012.
\newblock Large-scale item categorization for e-commerce.
\newblock In {\em Proceedings of the 21st ACM International Conference on
  Information and Knowledge Management}, CIKM '12, pages 595--604.

\bibitem[\protect\citename{Socher \bgroup et al.\egroup
  }2013]{socher2013recursive}
Richard Socher, Alex Perelygin, Jean~Y Wu, Jason Chuang, Christopher~D Manning,
  Andrew~Y Ng, and Christopher Potts.
\newblock 2013.
\newblock Recursive deep models for semantic compositionality over a sentiment
  treebank.
\newblock In {\em Proceedings of the conference on empirical methods in natural
  language processing (EMNLP)}, volume 1631, page 1642. Citeseer.

\bibitem[\protect\citename{Wang~Ling}2015]{wang2015}
Chris~Dyer Wang~Ling.
\newblock 2015.
\newblock Two/too simple adaptations of wordvec for syntax problems.
\newblock In {\em Proceedings of the 50th Annual Meeting of the North American
  Association for Computational Linguistics}. North American Association for
  Computational Linguistics.

\bibitem[\protect\citename{Xue \bgroup et al.\egroup
  }2008]{Xue2008deepclassification}
Gui-Rong Xue, Dikan Xing, Qiang Yang, and Yong Yu.
\newblock 2008.
\newblock Deep classification in large-scale text hierarchies.
\newblock In {\em Proceedings of the 31st Annual International ACM SIGIR
  Conference on Research and Development in Information Retrieval}, SIGIR '08,
  pages 619--626.

\end{thebibliography}
\bibliographystyle{emnlp2016}

\cleardoublepage

\section{Supplementary Material}

\subsection{Label Embedding Based Approach}
\label{sup_section:label_embedd}
Apart from tree based approaches there are label based embedding approches for Product Classification. Wei and Kwok \cite{Bi11multi-labelclassification} suggested a label based embedding approach which exploits label dependency in tree-structured hierarchies for hierarchical classification. Kernel Dependency Estimation (KDE) is used to first project or embed the label vector (multi-label) in fewer orthogonal dimensions. An advantage of this approach is that all m learners in the projected space can learn from the full training data. In contrast n tree based methods training data reduces as we reach leaf nodes.

To preserve dependencies during prediction the authors suggest a greedy approach. The problem can be efficiently solved using a greedy algorithm called Condensing Sort and Select Algorithm. However, the algorithm is computationally
intensive.

\subsubsection{Dependency Based Word Vectors}
\begin{figure}
\centering
\includegraphics[scale=0.3]{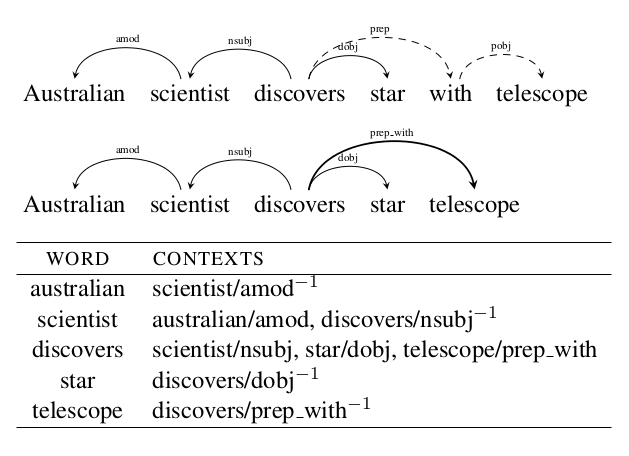}
\caption{Example for Dependency-based context extraction}
\label{dependency}
\end{figure}

SGNS and CBoW both use linear bag of words context for training word vectors \cite{mikolov2013distributed}. Levy and Goldberg \cite{levy2014dependencybased} suggested use of arbitrary functional context instead like syntactic dependencies generated from a parse of the sentence. Each word $w$ and its modifiers $m_{1},\ldots,m_{k}$ are extracted from a sentence parse. 
Contexts in the form ($m_{1}, lbl_{1}, \ldots, m_{k}$, $lbl_{k}$ ) are generated for every sentence. Here $lbl$ is the dependency relationship type between word and the modifier and $lbl^{-1}$ is used to denote the inverse relationship. Figure \ref{dependency} shows dependency based context for words in a given sentence.

The dependency based word vectors use the same training methods as SGNS.  Compared to similarly learned linear context based vectors learned it is found that the dependency based vectors perform better on functional similarity. However, for the task of topical similarity estimation the linear context based word vectors encode better distributional semantic content. 

\begin{algorithm}
    \KwData{Catalog Taxonomy Tree (T) of depth K and testing data $D$ = (d,$p_{d}$) where d is product description $p_{d}$ is taxonomy of paths. Set of level  one Classifiers C = \{{$PP$,$NP$,$DNP_{1}$ $\ldots$ $DNP_{K}$\}} and final level two classifier $FPP$}
    \KwResult{top m prediction path $P_{d_{i}}$ for training description d, here i = 1 $\ldots$ m}
    Obtain $\vec{gwBoWV_{d}}$ features for each product description d in test data\;
    Get Prediction Probabilities from all level one classifiers to obtain level two feature vector ($\vec{FV_{d}}$) using Equation \ref{fv}\;
    Obtain ($\vec{RFV_{d}}$) reduced feature vector\;
    Output top m paths from final prediction using output probabilities from level two  classifier $FFP$  for description d.
\caption{Testing Two Level Boosting Approach}
\label{test_ensemble}
\end{algorithm}

\subsection{Example of gwBoWV Approach}

\begin{enumerate}
    \item Assume there are four clusters $C = [ C_{1}, C_{2}, C_{3}, C_{4} ]$, here $C_{i}$ represents the $i^{th}$ cluster
    \item Let $D_{n} =  [ w_{1}, \ldots,w_{9}, w_{10}]$ be a document consisting of words $w_{1},w_{2},\ldots, w_{10}$ in order, whose document vectors need to be composed using word vectors $\vec{wv_{1}},\ldots,\vec{wv_{10}}$ respectively. Let us assume the following word-cluster assignment for document $D_{n}$ as given in Table \ref{CA:table}
    
    \begin{table}
    \centering
    \begin{tabular}{|l|l|}
    \hline
    \bf Word              & \bf Cluster       \\ \hline
    $w_{4},w_{3},w_{10},w_{5}$    & $C_{1}$ \\ \hline
    $w_{9}$            & $C_{2}$         \\ \hline
    $w_{1},w_{6},w_{2}$    & $C_{3}$ \\ \hline
    $w_{8},w_{7}$    & $C_{4}$ \\ \hline
    \end{tabular}
    \caption{\textit{Document Word Cluster Assignment}}
    \label{CA:table}
    \end{table}
    
    \item Obtain cluster $C_{i}$'s contribution in document $D_{n}$ by summation of word vectors for words coming from document $D_{n}$ and cluster $C_{i}$:
    
    \begin{itemize}
        \item $\vec{cv_{1}}$ = $\vec{wv_{4}}$+$\vec{wv_{3}}$ +$\vec{wv_{10}}$+$\vec{wv_{5}}$
        \item $\vec{cv_{2}}$ = $\vec{wv_{9}}$ 
        \item $\vec{cv_{3}}$ = $\vec{wv_{1}}$ + $\vec{wv_{6}}$+$\vec{wv_{2}}$
        \item $\vec{cv_{4}}$ = $\vec{wv_{8}}$+$\vec{wv_{7}}$
    \end{itemize}
    
    Similarly, calculate idf values for each cluster $C_{i}$ for document $D_{n}$:
    
    \begin{itemize}
        \item $icf_{1}$ = idf($w_{4}$)+idf($w_{3}$) +idf($w_{10}$)+idf($w_{5}$)
        \item $icf_{2}$ = idf($w_{9}$) 
        \item $icf_{3}$ = idf($w_{1}$) + idf($w_{6}$)+idf($w_{2}$)
        \item $icf_{4}$ = idf($w_{8}$)+idf($w_{7}$)
    \end{itemize}
    
    \item Concatenate cluster vectors to form Bag of Words Vector of dimension,  $\#cluster\times \#wordvec$:
    \begin{equation}
    \vec{BoWV(D_{n})} = \vec{cv_{1}}\oplus \vec{cv_{2}}\oplus \vec{cv_{3}}\oplus \vec{cv_{4}}
    \end{equation}

    \item Concatenate word-cluster idf values to form graded weighted Bag of Word Vector of dimension $\#cluster\times \#wordvec + \#cluster$:
    \begin{equation}
    \begin{split}
          \vec{gwBoWV(D_{n})} = \vec{cv_{1}}\oplus \vec{cv_{2}}\oplus 
          \vec{cv_{3}}\oplus \vec{cf_{4}} \oplus \\ icf_{1}\oplus icf_{2}\oplus icf_{3}\oplus icf_{4}. 
    \end{split}
    \end{equation}
\end{enumerate}

\subsection{Quality of WordVec Clusters}
Below are examples of words contained in some clusters and their possible cluster topic meaning for the book data. Each cluster is formed using clustering of word-vectors where the word belongs to particular topics. We number clusters according to the distance of the centroid from the origin to avoid confusion. \\

\begin{enumerate}
\item Cluster \#0  basically talks about crime and punishment related terms like \textit{“accused, arrest, assault, attempted, beaten, attorney,brutal,confessions, convicted cops, corrupt, custody, dealer, gang, investigative, gangster, guns, hated, jails, judge, mob, undercover, trail, police,  prison, lawyer, torture, witness etc“}
\item Cluster \#10 talks about scientific experiments related terms like \textit{“yield, valid, variance, alternatives, analyses, calculating, comparing, assumptions, criteria, determining, descriptive, evaluation, formulation, experiments, measures model, parameters, inference, hypothesis etc“}
\end{enumerate}

Similarly, Cluster \#13 is talking about dating and marriage, Cluster \#11 about tools and tutorials and Cluster \#15 about persons. Other clusters also represent single or multiple topics similiar to each other. Similarity of words within a cluster represents efficient distributional semantic representation of wordvectors trained by the SGNS model.

\subsection{Two Level Classification Approach}
We also experimented with a modified approach of  two level classification given by Shen and Ruvini \cite{Shen2011categorization} \cite{Shen2012categorization} as describe in Section \ref{treebased}. However, instead of randomly giving direction and then finding a dense graph using Strongly Connected Components, we decided the edge direction from misclassification and used various methods like weakly connected component, bi connected component and articulation points to find Highly Connected Component. We followed this approach to improve sensitivity and cover missing edges as discussed in section \ref{treebased}. The value of the confusion probability and direction of edges is decided by the value of $(i,j)$ element in the confusion matrix $(CM)$.

\begin{algorithm}
\label{mccg}
    \SetAlgoLined
    \KwData{Set of Categories $C = \{{c_{1},c_{2},c_{3} ...c_{n}}\}$ and threshold $\alpha$}
    \KwResult{ Set of dense sub-graphs $CG = \{{cg_{1},cg_{2},cg_{3},cg_{4} \ldots cg_{m}}\}$ representing highly connected groups}
    Train a weak classifier H on all possible categories \;
    Compute pairwise confusion probabilities between classes using values from the confusion matrix \textit{(CM)}.
    \begin{equation}
    Conf(c_{i},c_{j}) =  
        \begin{dcases}
        CM(c_{i},c_{j}),& \text{if }  CM(c_{i},c_{j}) \geq \alpha \\
        0              & \text{otherwise}
        \end{dcases}
    \end{equation}
    here, Conf($c_{i},c_{j}$) may not be equal to Conf($c_{j},c_{i}$) due to non symmetric nature of Confusion Matrix $CM$.\;
    Construct confusion graph $G=(E,V)$ with vertices ($V$) as confused categories and edges ($E_{ij}$) from $i\rightarrow j$ with weight = Conf($c_{i},c_{j}$).\;
    Apply Bi-Connected Component, Strongly Connected Component or Weakly Connected Component finding graph algorithm on $G$ to obtain set of dense sub-graphs $CG = \{{cg_{1},cg_{2},cg_{3},cg_{4} \ldots cg_{m}}\}$.

\caption{Modified Connected Component Grouping}
\end{algorithm}

\subsection{Confused Category Group Discovery}

\begin{figure}
\centering
\includegraphics[scale=0.25]{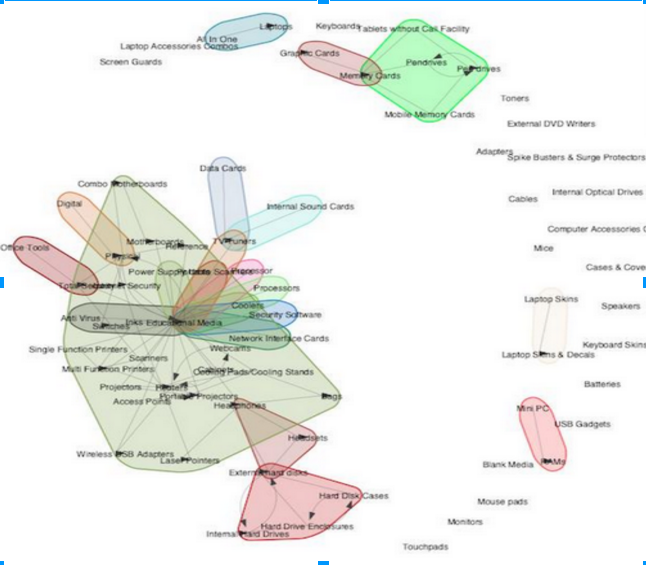}
\caption{Misclassification Graphs with latent groups in Computer and Computer Accessories, here each edge from C1 $\rightarrow$ C2 represents misclassification from C1 $\rightarrow$ C2 with threshold $> 10\%$ of correct prediction, here isolated vertexes represent almost correctly predicted classes}
\label{fig:mc1}
\end{figure}

Figure \ref{fig:mc1} shows \textit{“Hard Disk, Hard Drive, Hard Disk Case and Hard Drive Enclosure”} are misclassified as each other and form  a latent group in Computer and Computer accessories  extracted by finding bi-connected components in the misclassification graph.

\begin{figure}
\centering
\includegraphics[scale=0.3]{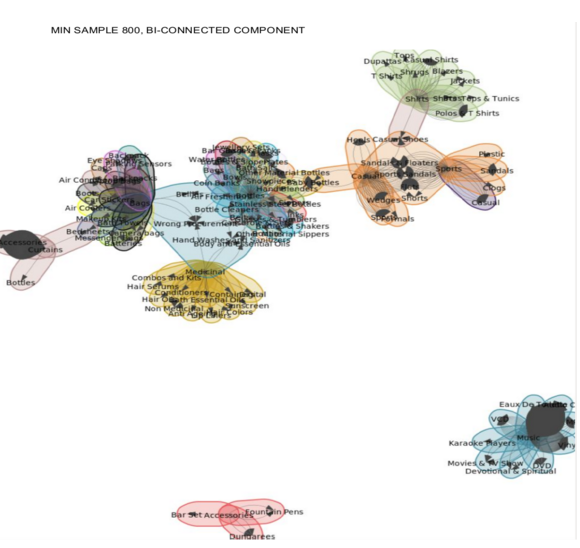}
\caption{Final Misclassification Graphs and latent groups (color bubble) discovered during search phase on multiple categories with threshold of 800 (weighted) examples for weakly bi-connected component.}
\label{fig:mc3}
\end{figure}

Figure \ref{fig:mc3} shows the final latent groups discovered (color bubble) in Non-Book Data using graded weighted Bag of Word Vector methods and random forest classifier without class balance on raw data with varying thresholds on \#mis classification for dropping edges based on edge weight.


\begin{figure}
\centering
\includegraphics[scale=0.25]{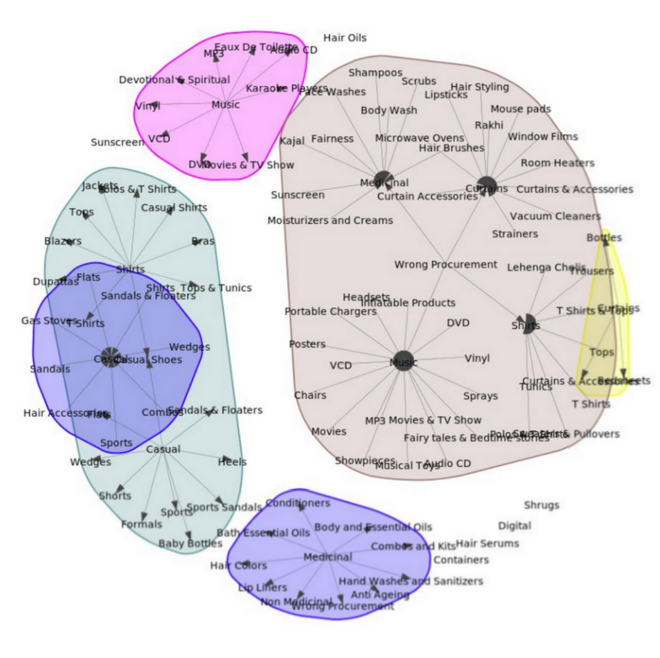}
\caption{Final Misclassification Graphs and latent groups (color bubble) discovered  during search phase on multiple categories with threshold of 1000 (weighted) example and above weakly bi-connected component. Here we represent multiple edge by single edge for image clarity.}
\label{5.2}
\end{figure}
\subsection{Data Visualization Tree Maps}

\begin{figure}
\centering
\includegraphics[scale=0.15]{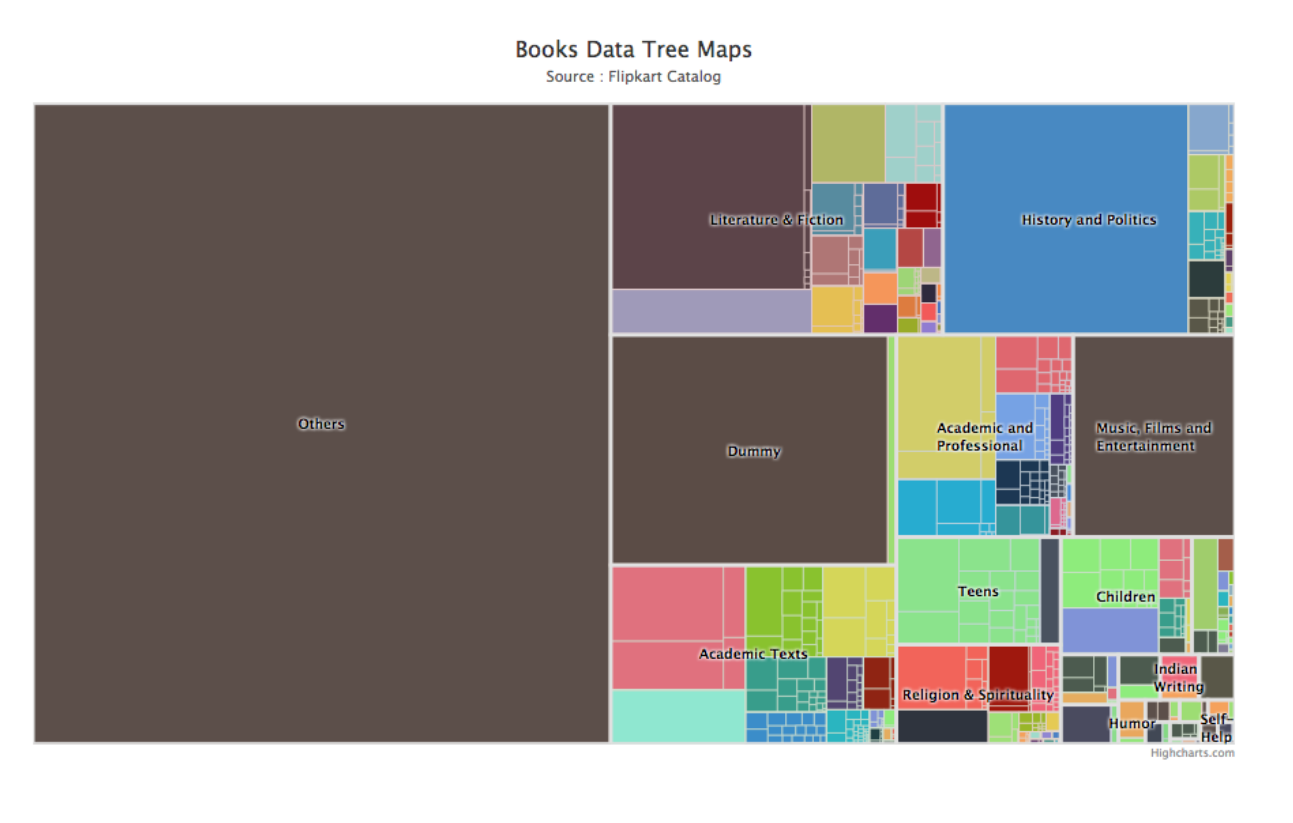}
\caption{ Book data visualization using Tree Map at root }
\label{tm1}
\end{figure}

\begin{figure}
\centering
\includegraphics[scale=0.15]{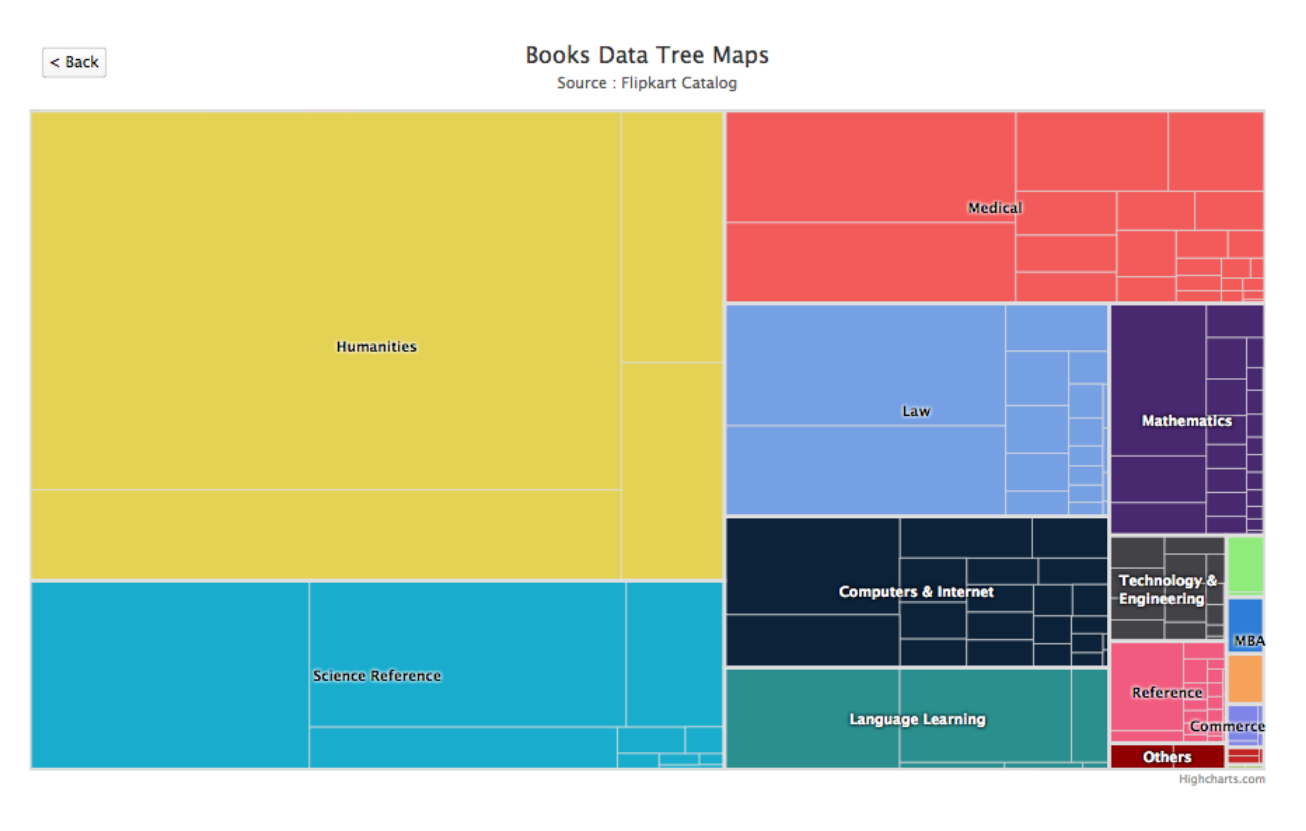}
\caption{ Book data visualization using Tree Map at depth 1 for Academic and Professional}
\label{tm2}
\end{figure}

We visualize the taxonomy for some main categories using Tree-Maps. Figures \ref{tm1} - \ref{tm2} show tree maps at various depths for the book taxonomy. It is evident from these maps that the tree exhibits high skewness and heavy tailed nature.

\begin{figure}
\centering
\includegraphics[scale=0.15]{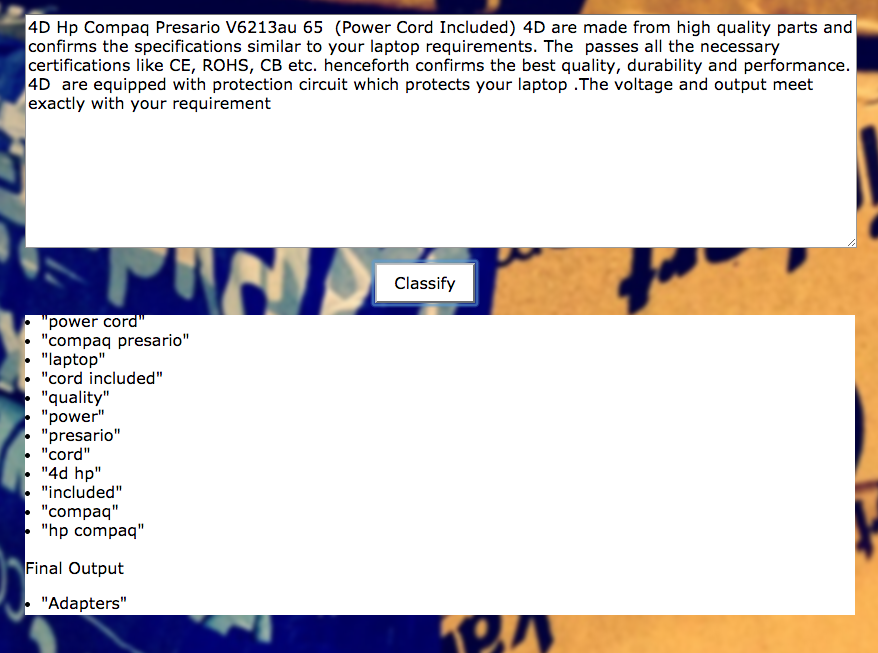}
\caption{Example result on Non Book Data, input description and output labels with top $k$ categories}
\label{5.16}
\end{figure}

\begin{figure}
\centering
\includegraphics[scale=0.15]{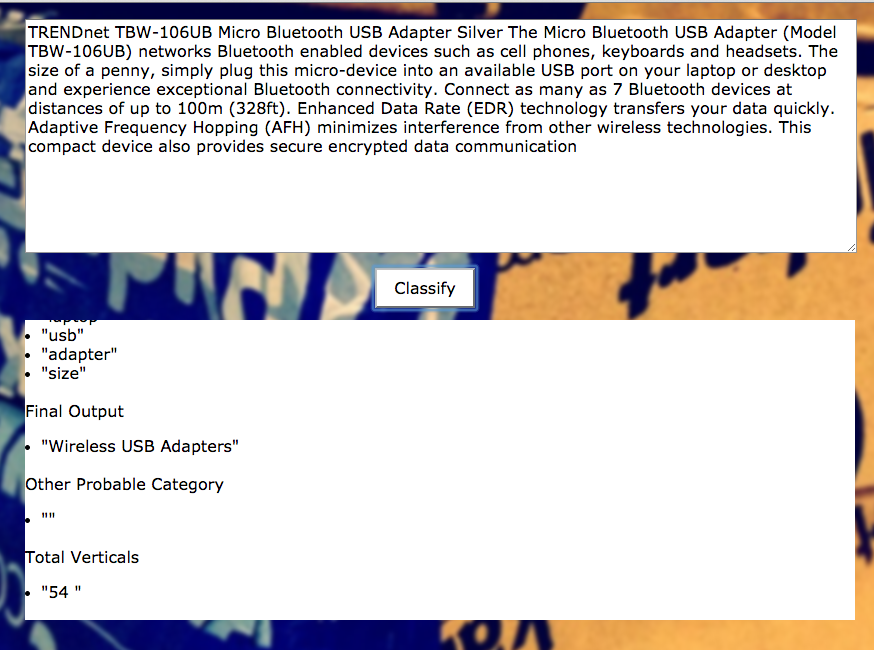}
\caption{Example result on Non Book Data, input description and output labels with top $k$ categories}
\label{5.17}
\end{figure}

\begin{figure}
\centering
\includegraphics[scale=0.1]{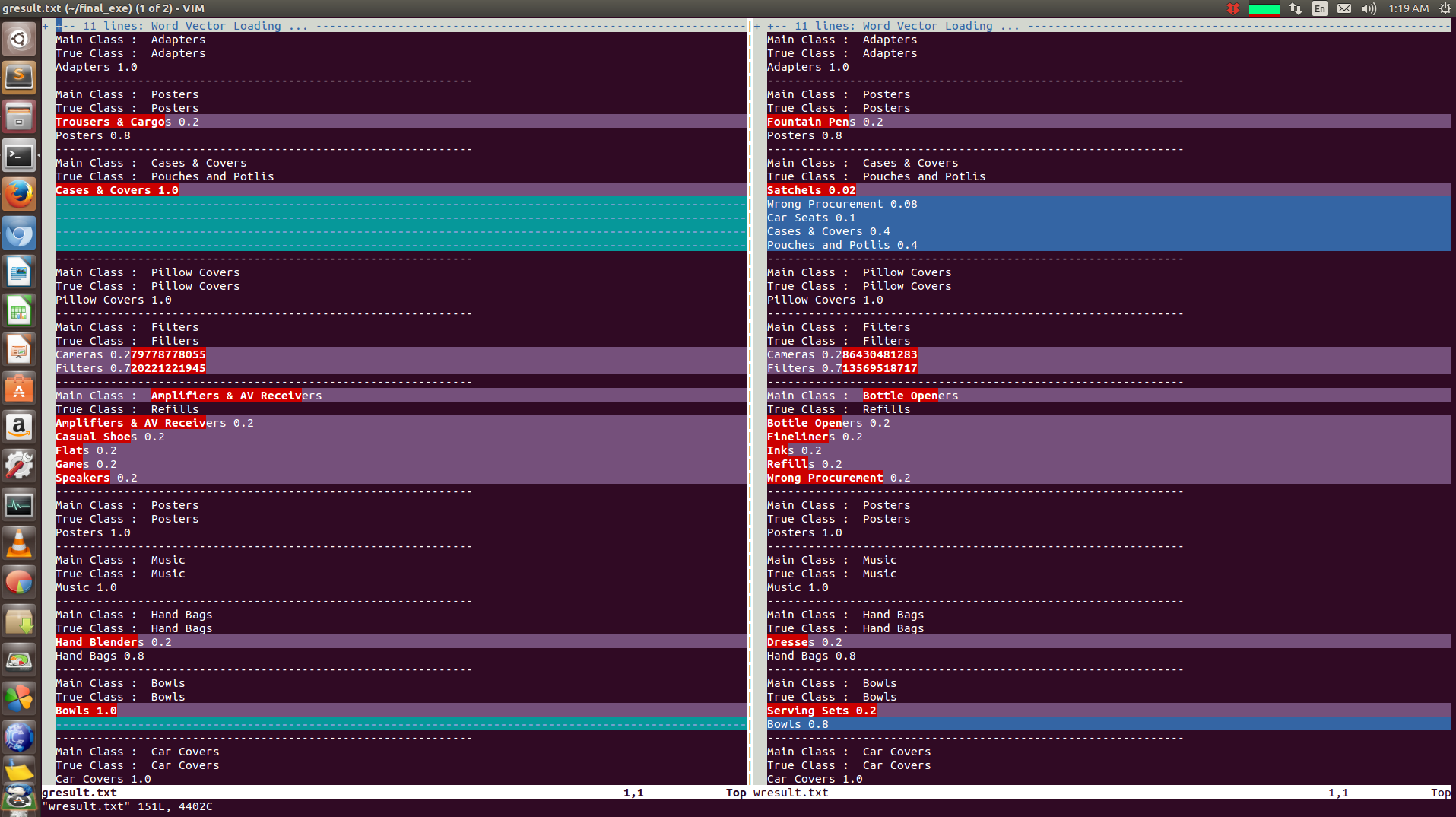}
\caption{ Result of Glove vs WordVec on Non Book Dataset}
\label{5.13}
\end{figure}


\subsection{More Results for Ensemble Approach}
We use kNN and random forest for initial classification instead of SVM because of better stability to class imbalance and better performance due to generation of good set of meta features. Also SVM doesn't perform well with huge number of classes. Table \ref{Table:5.2} confirm the same empirically

\begin{table}
\centering
\begin{tabular}{|l|l|}
\hline
\bf Algorithm  & \bf \%Accuracy \\ \hline
kNN       & 74.2\%  \\ \hline
multiclass svm       & 77.4\%  \\ \hline
random forest     &  \bf 79.6\%  \\ \hline
\end{tabular}
\caption{\textit{Comparison of flat classification with multiple classifiers using 95000 training, 9000 testing samples and 460 probable classes on non-book data-set using tf-idf features for Computer dataset }}
\label{Table:5.2}
\end{table}

\begin{table}
\centering
\begin{tabular}{|l|l|l|l|l|}
\hline
\bf Data & \bf\#Class &\bf \#Train & \bf\#Test &\bf \%CP@1 \\ \hline
Computer  & 54 & 0.1 & 2.5 & 93.0\% \\ \hline
Computer  & 54 & 0.66 & 3.3 & 98.5\% \\ \hline
Home  & 49 & 0.1 & 2.5 & 97.2\% \\ \hline
8-top cat & 460 & 0.09 & 0.95 & 77.3\% \\ \hline
8-top cat* & 460 & 0.09 & 0.95 & 79.0\% \\ \hline
20-top cat & 900 & 0.09 & 0.95 & 74.2\% \\ \hline 

\end{tabular}
\caption{\textit{Performance of flat classification using kNN classifier on various sample non-book categories(cat)
*Used ensemble here i.e. a  level two classifier trained on output probabilities of flat weak classifiers at level one. }}
\label{Table:5.3}
\end{table}

We observed improvement in classification accuracy by using Shen and Lee approach of two level classifier for discovering latent groups and running fine classifiers on them. Table \ref{Table:5.4} shows improvement in accuracy by a level two classifier.

\begin{table}
\centering
\begin{tabular}{|l|l|}
\hline
\bf Algorithm  & \bf \%Accuracy \\ \hline
kNN-SVM       & \bf 91.0\%  \\ \hline
kNN       & 78.0\%  \\ \hline
\end{tabular}
\caption{\textit{Coarse - fine level classification results on highly connected category set \{Hard Disk Cases, Hard Drive Enclosures,Internal Hard Disk and External Hard Drive\}  with 1457 training and 728 testing samples}}
\label{Table:5.4}
\end{table}

To prove that books were more confusing compared to non-book, we did a small experiment. We sampled all computer and computer accessories and Computer related books and binary labeled them and compared this with a direct classifier without binary labelling. The results are in table \ref{Table:5.17}.

\begin{table}
\centering
\begin{tabular}{|l|l|l|}
\hline
\bf Model & \bf \#Classes \bf & \bf \#Accuracy\\ \hline
Binary & 2 & 97\% \\ \hline
Non-Binary & 700 & 73\% \\ \hline

\end{tabular}
\caption{\textit{Accuracy drop due to misclassification within book categories on \#Training = 25000 and \#Testing = 10000}}
\label{Table:5.17}
\end{table}

\begin{table}
\centering
\begin{tabular}{|l|l|l|l|l|}
\hline
\bf Red-Dim & \bf \%PP & \bf \%CP & \bf \%LP & \bf \%LC\\ \hline
1000 & 40.30 & 74.45 & 86.38 & 22.47\\ \hline
2000 & 40.88  & 74.92  & 86.87 & 22.56\\ \hline
3000 & 40.99 & 75.24 & 87.06 & 22.60\\ \hline
4000 & 41.11  & 75.24  & 87.07  & 22.53\\ \hline
\end{tabular}
\caption{\textit{Results from reduced gwBoWV vectors for Top 6 path prediction (Orignal Dimension: 8080)}}
\label{Table 5.20}
\end{table}

\begin{table}
\centering
\begin{tabular}{|l|l|l|l|l|}
\hline
\bf Red-Dim & \bf \%PP & \bf \%CP & \bf \%LP & \bf \%LC\\ \hline
1000 & 46.26 & 72.46 & 84.84 & 24.85\\ \hline
2000 & 47.05 &.72.26 & 84.52 & 25.05\\ \hline
2500 &.47.70 & 72.77 & 84.58 & 24.81\\ \hline
3000 &.44.45 & 73.84 & 85.83 & 23.74\\ \hline
\end{tabular}
\caption{\textit{Results of varying reduced dimension for level one node classifer where level two classifier uses Top 6 prediction}}
\label{Table 5.23}
\end{table}

Results from various approaches on top 3 taxonomy prediction are in Table \ref{Table:8}. \ref{Table 5.23} shows results  of node level two classifier on various reduced dimension vectors (using ANOVA) - original vectors were concatenated output probabilities of node prediction probabilities using gwBoWV. \ref{Table 5.23} show results of level one classifier on various reduced dimension vectors (using ANOVA) where original vectors were gwBoWV. Ensemble and gwBoWV perform better than other approaches.


\begin{table}
\centering
\begin{tabular}{|l|l|l|l|l|}
\hline
\bf Method & \bf PP & \bf CP & \bf LP & \bf LC\\ \hline
tfidf(4000-2-20) & 44.68 & 71.34 & 83.36 & 40.02\\ \hline
tfidf(8000-2-20) & 44.70 & 71.18 & 83.30 & 40.08\\ \hline
tfidf(10000-2-20) & 44.69 & 71.21 & 83.30 & 40.13\\ \hline
path-1(4100-15) & 42.67 & 70.46 & 82.78 & 37.85\\ \hline
path-1(8080-20) & 44.48 & 71.13 & 83.21 & 40.26\\ \hline
depth(7975-20) & 44.91 & 71.49 & 83.52 & 40.54\\ \hline
node(4100+2810) & 44.86 & 71.04 & 83.23 & 40.34\\ \hline
comb-2(8000) & 47.62 & \bf 73.01 & \bf 85.16 & \bf 41.69\\ \hline
comb-2(6000) & \bf 48.17 & 72.07 & 84.36 & 41.52\\ \hline
comb-2(10000) & 45.78 & 71.85 & 84.03 & 40.81\\ \hline
\end{tabular}
\caption{\textit{Results from various approaches for Top 3 prediction}}
\label{Table:8}
\end{table}

\subsection{Classification Example from Book Data}
\textbf{Description} :  \textit{ignorance is bliss or so hopes antoine the lead character in martin pages stinging satire how i became stupida modern day candide with a darwin award like sensibility a twenty five year old aramaic scholar antoine has had it with being brilliant and deeply self aware in todays culture so tortured is he by the depth of his perception and understanding of himself and the world around him that he vows to renounce his intelligence by any means necessary in order to become stupid enough to be a happy functioning member of society what follows is a dark and hilarious odyssey as antoine tries everything from alcoholism to stock trading in order to lighten the burden of his brain on his soul. how i became stupid. how i became stupid. how i became stupid} \\
\textbf{Actual Class} : books-tree $\rightarrow$ literature and fiction

\textbf{Predictions}, \textbf{Probability} \\

books-tree $
\rightarrow$ literature and fiction $
\rightarrow$ literary collections $\rightarrow$ essays 0.1 \\
books-tree $\rightarrow$ reference $\rightarrow$ bibliographies and indexes 0.1 \\
books-tree $\rightarrow$ hobbies and interests $\rightarrow$ travel $\rightarrow$ other books $\rightarrow$ reference 0.1 \\
books-tree $\rightarrow$ children $\rightarrow$ children literature $\rightarrow$ fairy tales and bedtime stories 0.1 \\
books-tree $\rightarrow$ dummy 0.2 \\

\textbf{Description} :  \textit{harpercollins continues with its commitment to reissue maurice sendaks most beloved works in hardcover by making available again this 1964 reprinting of an original fairytale by frank r stockton as illustrated by the incomparable maurice sendak in the ancient country of orn there lived an old man who was called the beeman because his whole time was spent in the company of bees one day a junior sorcerer stopped at the hut of the beeman the junior sorcerer told the beeman that he has been transformed if you will find out what you have been transformed from i will see that you are made all right again said the sorcerer could it have been a giant or a powerful prince or some gorgeous being whom the magicians or the fairies wish to punish the beeman sets out to discover his original form. the beeman of orn. the beeman of orn. the beeman of orn.} \\
\textbf{Actual Class} :  books-tree $\rightarrow$ children $\rightarrow$ knowledge and learning $\rightarrow$ animals books $\rightarrow$ reptiles and amphibians

\textbf{Predictions}, \textbf{Probability} \\

books-tree $\rightarrow$ children $\rightarrow$ knowledge and learning $\rightarrow$ animals books $\rightarrow$ reptiles and amphibians , 0.28 \\
books-tree $\rightarrow$ children $\rightarrow$ fun and humor, 0.72 \\

\textbf{Description} :  \textit{a new york times science reporter makes a startling new case that religion has an evolutionary basis  for the last 50000 years and probably much longer people have practiced religion yet little attention has been given to the question of whether this universal human behavior might have been implanted in human nature in this original and thought provoking work nicholas wade traces how religion grew to be so essential to early societies in their struggle for survival how an instinct for faith became hardwired into human nature and how it provided an impetus for law and government the faith instinct offers an objective and non polemical exploration of humanity’s quest for spiritual transcendence. the faith instinct how religion evolved and why it endures. the faith instinct how religion evolved and why it endures. the faith instinct how religion evolved and why it endures}\\
\textbf{Actual Class} :  books-tree $\rightarrow$ academic texts $\rightarrow$ humanities

\textbf{Predictions}, \textbf{Probability} \\
books-tree $\rightarrow$ academic texts $\rightarrow$ humanities 0.067 \\
books-tree $\rightarrow$ religion and spirituality $\rightarrow$ new age and occult $\rightarrow$ witchcraft and wicca 0.1 \\
books-tree $\rightarrow$ health and fitness $\rightarrow$ diet and nutrition $\rightarrow$ diets 0.1 \\
books-tree $\rightarrow$ dummy 0.4 \\

\textbf{Description} :  \textit{behavioral economist and new york times bestselling author of predictably irrational dan ariely returns to offer a much needed take on the irrational decisions that influence our dating lives our workplace experiences and our general behaviour up close and personal  in the upside of irrationality behavioral economist dan ariely will explore the many ways in which our behaviour often leads us astray in terms of our romantic relationships our experiences in the workplace and our temptations to cheat blending everyday experience with groundbreaking research ariely explains how expectations emotions social norms and other invisible seemingly illogical forces skew our reasoning abilities  among the topics dan explores are  what we think will make us happy and what really makes us happy  why learning more about people make us like them less  how we fall in love with our ideas  what motivates us to cheat  dan will emphasize the important role that irrationality plays in our daytoday decision making not just in our financial marketplace but in the most hidden aspects of our livesabout the author an ariely is the new york times bestselling author of predictably irrational over the years he has won numerous scientific awards and his work has been featured in leading scholarly journals in psychology economics neuroscience and in a variety of popular media outlets including the new york times the wall street journal the washington post the new yorker scientific american and science. the upside of irrationality. the upside of irrationality. the upside of irrationality}\\
\textbf{Actual Class} :  books-tree $\rightarrow$ business, investing and management $\rightarrow$ business $\rightarrow$  economics

\textbf{Predictions}, \textbf{Probability}
books-tree $\rightarrow$ business, investing and management $\rightarrow$ business $\rightarrow$ economics 0.15\\
books-tree $\rightarrow$ philosophy $\rightarrow$ logic 0.175\\
books-tree $\rightarrow$ self-help $\rightarrow$ personal growth 0.21\\
books-tree $\rightarrow$ academic texts $\rightarrow$ mathematics 0.465 \\
\end{document}